# Exponential Families for Conditional Random Fields

**Yasemin Altun**[*]  
[*]Department of Computer Science  
Brown University  
Providence, RI 02912, USA

**Alex J. Smola**[†]  
[†]Machine Learning Program  
National ICT Australia and ANU  
Canberra 0200, ACT, Australia

**Thomas Hofmann**[*‡]  
[‡]Max-Planck Institut for  
Biological Cybernetics  
Tübingen, Germany

## Abstract

In this paper we define conditional random fields in reproducing kernel Hilbert spaces and show connections to Gaussian Process classification. More specifically, we prove decomposition results for undirected graphical models and we give constructions for kernels. Finally we present efficient means of solving the optimization problem using reduced rank decompositions and we show how stationarity can be exploited efficiently in the optimization process.

## 1 Introduction

The benefits of a framework for designing flexible and powerful input representations for machine learning problems has been demonstrated impressively by the success of kernel-based methods. However, many real-world prediction problems also involve complex output spaces, with possible dependencies between multiple output variables. Among the many examples are classification problems with a Markov-chain dependency structure, which is typical in many natural language and speech processing tasks (e.g. part-of-speech tagging, named entity classification, shallow parsing, pitch accent prediction), but which is also relevant for tasks like optical character recognition, text to phoneme transcription, secondary protein structure prediction and many other problems. More complicated dependency structures are also commonplace, the simplest example being multi-label classification.

A well-known approach for solving these problems are Conditional Random Fields (CRFs), proposed by Lafferty et al. [2001], an extension of logistic regression that can take dependencies between labels in a graph ( most commonly between neighboring labels along a chain) into account. Related approaches include the work of Punyakonok and Roth [2001] and McCallum et al. [2000]. More recently, Altun et al. [2003] and Taskar et al. [2003] have presented similar extensions of multiclass Support Vector Machine (SVM) learning. In this paper, we provide further theoretical underpinnings for the work of Lafferty et al. [2001] and Taskar et al. [2003] by investigating the relationship between kernelized exponential families and graphical models. On the practical side, this leads to the derivation of an efficient estimation algorithm for kernelized CRFs.

## 2 Exponential Families

### 2.1 Definition and Basic Facts

Exponential families are a basic engine for estimation in the present paper. An exponential family is a $\theta$-parameterized family of probability density functions $p(z|\theta)$ that can be written in canonical form as follows:

$$p(z|\theta) = p_0(z) \exp\left(\langle \Phi(z), \theta \rangle - g(\theta)\right). \qquad (1)$$

Here $\theta$ is the canonical or natural parameter, $\Phi(z)$ is the corresponding vector of sufficient statistics and $g(\theta)$ is the log-partition function or moment generating function,

$$g(\theta) = \log \int_{\mathcal{Z}} p_0(z) \exp(\langle \Phi(z), \theta \rangle) dz, \qquad (2)$$

where $\mathcal{Z}$ is the domain of $z$. Moreover, $\langle \cdot, \cdot \rangle$ denotes the scalar product in a Euclidean space or – as we will require at a later stage – a scalar product $\langle \cdot, \cdot \rangle_{\mathcal{H}}$ in a Reproducing Kernel Hilbert space (RKHS) $\mathcal{H}$. Finally, $p_0(z)$ is a probability mass function or probability density that can be used to model a change of measure. Typically $p_0$ is set to be constant.

The log-partition function $g(\theta)$ plays an important role in estimation. In particular, it can be used to compute the moments of the distribution, see e.g. Lauritzen [1996]:

**Proposition 1** *The log-partition function $g(\theta)$ is a convex $C^\infty$ function. Moreover, the derivatives of $g$ generate the corresponding moments of $\Phi(z)$, i.e.*

$$\partial_\theta g(\theta) = \mathbf{E}_{p(z|\theta)}[\Phi(z)] \quad \text{Mean} \quad (3a)$$
$$\partial_\theta^2 g(\theta) = \text{Var}_{p(z|\theta)}[\Phi(z)] \quad \text{Variance}. \quad (3b)$$

Since our main interest is in cases, where a fixed set of covariates $x$ is given as the input and the goal is to predict a – possibly structured and complex – response variable $y$, we may use (1) with $z = (x, y)$ to determine the functional form of conditional distributions $p(y|x;\theta)$ simply via

$$p(y|x;\theta) = \exp\left(\langle\Phi(x,y),\theta\rangle - g(\theta;x)\right), \quad (4)$$

Here for given $x$, $\Phi(x,y)$ is a vector of sufficient statistics of $p(y|x)$ and $g(\theta;x)$ is the conditional log partition function

$$g(\theta;x) := \sum_{y \in \mathcal{Y}} \exp(\langle\Phi(x,y),\theta\rangle), \quad (5)$$

where we have assumed a discrete space $\mathcal{Y}$. Analogous to Proposition 1 we have

**Proposition 2** *$g(\theta;x)$ is a convex $C^\infty$ function. Moreover, the derivatives of $g$ satisfy*

$$\partial_\theta g(\theta;x) = \mathbf{E}_{p(y|x;\theta)}[\Phi(x,y)|x] \quad \text{Mean} \quad (6a)$$
$$\partial_\theta^2 g(\theta;x) = \text{Var}_{p(y|x;\theta)}[\Phi(x,y)|x] \quad \text{Variance}. \quad (6b)$$

### 2.2 Universal Density Estimators

We now provide a motivation for using exponential families with rich sufficient statistics. One can show that if $\Phi(z)$ is powerful enough, exponential families become universal density estimators. This is advantageous, as it opens the domain of nonparametric estimation to an area of statistics which so-far was restricted to parametric distributions.

In the following we will be working in a RKHS and it is more general to use $f$ defined via

$$f(z) = \langle f, k(z,\cdot)\rangle_\mathcal{H}, \quad (7)$$

such that $f(\cdot) = \langle\Phi(\cdot),\theta\rangle$, in the case that $k(\cdot,\cdot) = \langle\Phi(\cdot),\Phi(\cdot)\rangle$, and we can use $f$ and $\theta$ synonymously.

**Proposition 3 (Dense Densities)** *Let $\mathcal{Z}$ be a measurable set with respect to the Lebesgue measure and denote by $\mathcal{P}$ a family of densities on $\mathcal{Z}$, for which the density is bounded from above and continuous. Furthermore, let $k$ be a kernel defined on $\mathcal{Z} \times \mathcal{Z}$ with corresponding RKHS $\mathcal{H}$ which is dense in the space of continuous functions on $\mathcal{Z}$, that is $C^0(\mathcal{Z})$, in the $L_\infty$ sense. Then the family of distributions defined in (1) are also dense in $\mathcal{P}$ in the $L_\infty$ sense.*

**Proof** Let $Z := \mu(\mathcal{Z})$ be the measure of $\mathcal{Z}$. Then for any $p \in \mathcal{P}$ let $C := \max_{z \in \mathcal{Z}} |\log p(z)|$. By the fact that $\mathcal{H}$ is dense in $C^\infty(\mathcal{Z})$ there exists for every $\epsilon > 0$ some $f \in \mathcal{H}$ such that for all $z \in \mathcal{Z}$,

$$|f(z) - \log p(z)| \leq \frac{\epsilon}{2+2Z}e^{-2C}.$$

The latter, however, yields $|\exp(f(z)) - p(z)| \leq \frac{\epsilon}{2+2Z}e^{-C}$, as $|f|, |\log p| \leq C$. This implies

$$\left|\int_\mathcal{Z} \exp(f(z))dz - 1\right| \leq \frac{\epsilon Z}{2+2Z}e^{-C}$$

and consequently the log-partition function $g(f)$ is bounded by $\frac{\epsilon Z}{1+Z}e^{-C}$. Finally $|f(z) - g(f) - \log p(z)| \leq \epsilon e^{-C}$. Exponentiating terms proves the claim. ∎

Many RKHS $\mathcal{H}$ are dense in $C^0(\mathcal{X})$. See Steinwart [2001] for examples and a proof. This shows that choosing a density from a suitable exponential family is not restrictive in terms of approximation quality. What Proposition 3 does *not* prove is a bound on the approximation rate for any specific class of densities or a specific class of kernels.

## 3 Kernels for Markov Networks

The second ingredient that we need in order to develop our framework are Markov networks. Central to the following section is the celebrated Hammersley-Clifford Theorem, which we state here in a form convenient for our purposes.

**Theorem 4 (Hammersley-Clifford)** *Let random variables $Z = \{Z_1, \ldots, Z_k\}$ have a joint probability density (or mass) function with full support. Then $Z$ is a Markov random field with respect to an undirected graph $G = (Z, E)$ if and only if $Z$ has a Gibbs distribution with respect to $G$. The latter means that the joint probability density function over $Z$ can be written as*

$$p(z) = \exp\left(\sum_{C \in \mathcal{C}} \psi_C(z_C)\right) \quad (8)$$

*where $z_C$ denotes the restriction of $z$ on the maximal cliques $C$ of $G$ and $\psi_C$ denotes real-valued functions defined on the maximal cliques.*

What can we say about the distribution of random variables that form a Markov random field with respect to a graph $G$ and that are at the same time member of an exponential family? It turns out that the Gibbs form translates into a simple decomposition of the sufficient statistics and in turn the kernel function.

**Lemma 5 (Decomposition of $\Phi$)** *For positive probability density functions over a Markov random field $Z$ on $G$, the sufficient statistics $\Phi(z)$ satisfy*

$$\Phi(z) = (\Phi_{C_1}(z_{C_1}), \ldots, \Phi_{C_n}(z_{C_n})), \quad (9)$$

*where $C_i$ are the maximal cliques of $G$. Moreover, kernels $k(z, z') = \langle \Phi(z), \Phi(z') \rangle$ satisfy*

$$k(z, z') = \sum_{C \in \mathcal{C}} k_C(z_C, z'_C). \quad (10)$$

**Proof** By construction we know that $\log p(z|\theta) = \langle \Phi(z), \theta \rangle - g(\theta; z)$ for all $z, \theta$. Furthermore, we also know by Theorem 4, that $\log p(z|\theta) = \sum_{C \in \mathcal{C}} \psi_C(z_C; \theta)$. In other words, there exist functions $\psi_C(z_C|\theta)$ such that

$$\langle \Phi(z), \theta \rangle = g(\theta; z) + \sum_{C \in \mathcal{C}} \psi_C(z_C; \theta), \ \forall \ z \in \mathcal{Z} \text{ and } \theta.$$

Since for any $\theta$ this needs to hold for all $z \in \mathcal{Z}$, we can pick an orthonormal basis of $\theta$, say $e_i$ and rewrite

$$\langle \Phi(z), e_i \rangle = \sum_{C \in \mathcal{C}} \eta_C^i(z_C)$$

for some scalar functions $\eta_C^i(z_C)$. The key point is that $\eta_C^i$ depends on $z$ only via its restriction on $z_C$. Next we set

$$\Phi_C(z_C) := (\eta_C^1(z_C), \eta_C^2(z_C), \ldots)$$

which allows us to compute

$$\langle \Phi(z), \theta \rangle = \left\langle \Phi(z), \sum_i e_i \theta_i \right\rangle = \sum_i \theta_i \langle \Phi(z), e_i \rangle$$
$$= \sum_{C \in \mathcal{C}} \sum_i \theta_i \eta_C^i(z_C).$$

Rearranging terms shows that $\Phi$ satisfies the claim. The second part of the claim follows from $k_C(z, z') = \langle \Phi_C(z), \Phi_C(z') \rangle$. ∎

Note that this is merely an *existence proof*. In other words, while the lemma tells us the overall structure of the kernel, it does not make any particular statement on the form of the kernel functions $k_C$.

The following lemma specifies the kernel expansion arising from the Representer Theorem in cases where kernels are given by sums of kernel functions on smaller cliques.

**Lemma 6 (Kernel Decomposition)** *Denote by $k : \mathcal{Z} \times \mathcal{Z}$ a kernel function which can be written as $k(z, z') = \sum_{C \in \mathcal{C}} k_C(z_C, z'_C)$, where $z_C$ denotes the restriction of $z$ onto a subset of its coordinates, let $S = \{z^1, \ldots, z^m\}$ denote a sample and let $\alpha \in \mathbb{R}^m$. Then the following decomposition holds:*

$$\sum_{i=1}^m \alpha_i k(z^i, z) = \sum_{C \in \mathcal{C}} \sum_{\omega \in C} \bar{\alpha}_C^\omega k_C(\omega, z_C) \quad (11)$$

*for some set of parameters $\bar{\alpha}$, and where the sums over $\omega \in C$ go over all configurations of the clique $C$.*

**Proof** Direct calculation by setting $\bar{\alpha}_C^\omega = \sum_{i: z_C^i = \omega} \alpha_i$. ∎

While this lemma borders on triviality it is actually rather powerful: assume that $S$ contains an (exponentially) large number of binary sequences and $C$ was the restriction of $z$ on adjacent pairs along a sequence. Then (11) tells us that instead of $m = |S|$ parameters we only need to store as many parameters as there are different configurations in each clique.

For instance, the above statement generalizes the decomposition of kernel functions by Taskar et al. [2003], by replacing their reasoning concerning the Lagrange multipliers by the simple insight that the kernel used decomposes orthogonally.

## 4 Conditional Random Fields

In Conditional Random Fields (CRFs), we assume that given covariates $x$, the dependencies between the output variables $y$ can be modeled by a $\theta$-parameterized conditional exponential family which is also Markov with respect to a graph $G$. We are interested in estimating the parameters $\theta$ of the CRF from training data $(X, Y) \equiv \{(x^i, y^i), i = 1, \ldots, m\}$ where each $(x^i, y^i)$ is an instantiation of the nodes of $G$.

The sufficient statistics $\Phi(x, y)$ are of central interest in the following. More specifically, we are interested in the scalar products

$$k((x, y), (x', y')) := \langle \Phi(x, y), \Phi(x', y') \rangle. \quad (12)$$

It turns out that in complete analogy to Gaussian Process classification, we can perform estimation simply by evaluating scalar products between sufficient statistics. For instance, if $\mathcal{Y}$ is finite and if the optimal estimate $\theta$ itself is a convex combination of $\Phi(x, y)$, we can compute (5) without the need for evaluating $\Phi(x, y)$ explicitly.

To elaborate on this, we take the following log-likelihood function as our starting point

$$l(\theta; X, Y) = \sum_{i=1}^n \left( \langle \Phi(x^i, y^i), \theta \rangle - g(\theta; x^i) \right). \quad (13)$$

In order to perform estimation for high-dimensional families, one needs a proper prior on $\theta$, as the maximum likelihood estimate will inevitably lead to overfitting and bad generalization performance. Johnson et al. [1999] use a normal distribution on $\theta$ for a specific set of features $\Phi(x, y)$. We simply extend this to the general case

$$\theta \sim \mathcal{N}(0, \sigma^2 \mathbf{1}). \tag{14}$$

**Remark 7 (Covariance Function)** *It follows directly from [Williams, 1999] that a normal prior on $\theta$ corresponds to a Gaussian Process on the scalar products $\langle \Phi(x,y), \theta \rangle$ with covariance function $\sigma^2 k((x,y),(x',y'))$ and zero mean.*

*Note that for $\mathcal{Y} = \{\pm 1\}$ and $\Phi(x, y) = y\Phi(x)$ we have $k((x,y),(x',y')) = yy'k(x,x')$, which is exactly the kernel used in binary SVM and GP classification. For $\Phi(x, y) = e_y \times \Phi(x)$ (where $e_y$ is an element of the canonical basis) we obtain the kernel proposed in [Williams and Barber, 1998]. This clearly shows that CRFs are just a generalization of GP classification.*

With the above prior distribution the negative log-posterior $p(\theta|X, Y)$ for a conditional random field is given by

$$-\log p(\theta|X, Y) = \frac{1}{2\sigma^2} \|\theta\|^2 - l(\theta; X, Y) + \text{const.} \tag{15}$$

We approximate the Bayesian solution by a point estimate for $\theta$, namely the Maximum a Posteriori (MAP) estimate which is obtained by minimizing (15). The Representer Theorem [Schölkopf et al., 2001] states that the minimizer of (15) can be written as a linear combination of sufficient statistics over training inputs with suitably chosen weights $\alpha_y^i$:

$$\theta^{\text{MAP}} = \sum_{i=1}^{m} \sum_{y \in \mathcal{Y}} \alpha_y^i \Phi(x^i, y) \tag{16}$$

Note the sum over $\mathcal{Y}$, which is due to the fact that the conditional log-partition function sums over all $y \in \mathcal{Y}$.

Instead of a gradient descent procedure we advocate in this paper that a Gauss-Newton method can be used efficiently for estimation. Moreover, in order to deal with the large number of parameters we use a sparse greedy decomposition of the solution space. We give details in Section 5. We need some more results on the decomposition of $k$ based on the results of Section 3 for an efficient formulation and approximate methods for inference in the case where conditional expectations are too expensive to compute.

First of all, we state the following corollary of Lemma 6:

**Corollary 8 (Subspace Representer Theorem)**
*Equation (11) holds for the minimizer of the negative log-posterior (15).*

Note that nowhere we used the fact that $k$ was actually derived from a graphical model. Instead, it holds for any kernel which decomposes into a set of simpler kernel functions. This insight allows us to find lower bounds on the value of the objective function of many convex optimization problems.

**Lemma 9 (Lower Bound on Convex Functions)**
*Let $C : \Theta \to \mathbb{R}$ be a convex function on a vector space, let $\lambda \geq 0$, $\theta_0 \in \Theta$ and denote by $g \in \partial_\theta C(\theta_0)$ a vector in the subdifferential of $C$ at $\theta_0$. Then*

$$\min_{\theta \in \Theta} C(\theta) + \frac{\lambda}{2} \|\theta\|^2 \geq C(\theta_0) + \frac{\lambda}{2} \|\theta_0\|^2 - \frac{1}{2\lambda} \|g + \lambda \theta_0\|^2. \tag{17}$$

**Proof** Since $C$ is convex, it follows that for any subdifferential $g \in \partial_\theta C(\theta_0)$ we have $C(\theta) \geq C(\theta_0) + g^\top \delta\theta$. Consequently

$$\min_{\theta \in \Theta} C(\theta) + \frac{\lambda}{2} \|\theta\|^2 \geq \min_{\delta\theta \in \Theta} C(\theta_0) + g^\top \delta\theta + \frac{\lambda}{2} \|\theta_0 + \delta\theta\|^2. \tag{18}$$

The minimum is obtained for $\delta\theta = -(\lambda^{-1}g + \theta_0)$, which proves the claim. ∎

Again, this result seems rather trivial. However, it provides a valuable selection and stopping criterion for the inclusion of subspaces when optimizing over $\theta$. Note in particular that $g + \lambda\theta_0$ is the gradient of the optimization problem in (17), hence we obtain a lower bound on the objective function in terms of the 2-norm of the gradient and the regularization parameter $\lambda$.

**Corollary 10 (Subspace Descent)** *Denote by $\Theta_\perp \oplus \Theta_\| = \Theta$ an orthogonal decomposition of the space of natural parameters $\theta \in \Theta$ with corresponding decomposition $\theta = \theta_\perp + \theta_\|$. Then for $\theta \in \Theta_\|$*

$$\partial_{\theta_\|} \left[ C(\theta) + \frac{\lambda}{2} \|\theta\|^2 \right] = 0 \quad \text{and} \quad \partial_{\theta_\perp} C(\theta) = g_\perp$$

*the improvement achievable by letting $\theta \in \Theta$ is bounded by $\frac{1}{2\lambda} \|g_\perp\|^2$.*

In the context of decomposing kernels as in (11) this means that optimization over a clique $C$ with associated $k_C$ is only useful if the gradient in the corresponding direction is large enough.

Moreover, descent over the subspaces spanned by each clique are much cheaper to compute, as this involves computing correlations only within each clique for computation of first-order gradients.

**Remark 11 (Learning Structure)** *Given a set of possible cliques $\mathcal{C}$, find a graphical model which fits the data well. This can be achieved by performing subspace descent only in those subspaces where the initial gradient is sufficiently large, as in the other subspaces we have an upper bound on the maximum improvement.*

We now proceed to a further lemma allowing us to remove some of the cliques when performing conditional estimation, i.e. whenever we estimate $p(y|x)$.

**Lemma 12 (Irrelevant Cliques)** *Denote by $G$ an undirected graph on the pair of random variables $(x, y)$. Furthermore let $\mathcal{C} = \mathcal{C}_{\neg y} \cup \mathcal{C}_y$, where $\mathcal{C}$ is the set of all maximal cliques, $\mathcal{C}_{\neg y}$ is the set of all cliques for which the restrictions of $(x, y)$ are solely contained in $x$ and let $\mathcal{C}_y$ be its complement.*

*Then $p(y|x)$ only depends on $\Phi_C((x, y)_C)$ where $C \in \mathcal{C}_y$, that is, it suffices to study the kernel*

$$k((x,y),(x',y')) = \sum_{C \in \mathcal{C}_y} k_C((x,y)_C, (x',y')_C).$$

**Proof** We decompose $\Phi(x, y)$ into terms $\Phi_{\neg y}(x)$ dependent on $x$ alone and the remainder $\Phi_y(x, y)$ which contains all $\Phi_C(x, y)$ for which $C \in \mathcal{C}_y$ and likewise $\theta = (\theta_{\neg y}, \theta_y)$. Since $p(y|x, \theta) = \frac{p(y,x|\theta)}{\sum_{y'} p(y', x|\theta)}$ the conditional probability is independent of $\langle \Phi_{\neg y}(x), \theta_{\neg y} \rangle$. Consequently we can drop it, thus proving our claim. ∎

This reasoning gives us the kernel decomposition indicated implicitly by Lafferty et al. [2001], Sha and Pereira [2003] in the design of CRFs.

Again, the proof of the claim was relatively straightforward, yet the implications are rather deep: the key difference is that while in conditional random fields the assumption of specific functions on cliques is made in order to obtain the desired algorithms, in the above case it is a *consequence* of the fact that features on the maximal cliques containing only variables from $x$ are irrelevant for estimation.

## 5 Optimization for CRFs

We begin by describing a simple optimization strategy which is commonly used for small sample size estimates in Gaussian Processes and adapt it to CRFs. Subsequently, we give a modification of the second order method to a block-diagonal preconditioning and subspace descent to deal with the issue of computing conditional covariances over distant cliques. Finally we will show how low-rank decompositions can be used to reduce the number of variables involved to find a solution which is optimal in a subspace. It is computationally advantageous also for the purpose of classification of new observations.

### 5.1 Second Order Methods

It is well known [Fletcher, 1989] that for twice differentiable functions $L(\theta)$ the Newton updates

$$\theta \leftarrow \theta - \left[\partial_\theta^2 L(\theta)\right]^{-1} \partial_\theta L(\theta) \quad (19)$$

are quadratically convergent in a neighborhood of the minimizer of $L$. The chain rule for differentiation yields that for the parameterization $\theta = A\alpha$ we obtain the following update rule for $\alpha$:

$$\alpha \leftarrow \alpha - \left[A^\top \partial_\theta^2 L(\theta) A\right]^{-1} \left[\partial_\theta L(\theta) A\right] \quad (20)$$

Moreover, for convex functions the Newton method converges to the minimum, provided that convergence occurs. We now compute gradient and Hessian of the negative log-posterior $\mathcal{P} := -\log p(\theta|X, Y)$ for optimization.

$$\partial_\theta \mathcal{P} = \sum_{i=1}^m -\Phi(x^i, y^i) + \mathbf{E}_y\left[\Phi(x^i, y)|\theta, x^i\right] + \sigma^{-2}\theta \quad (21)$$

$$\partial_\theta^2 \mathcal{P} = \sum_{i=1}^m \mathrm{Cov}_y\left[\Phi(x^i, y)|\theta, x^i\right] + \sigma^{-2}\mathbf{1} \quad (22)$$

As the sufficient statistics may only be given implicitly, we can evaluate (21) and (22) only via scalar products, that is $\langle \Phi(x, y), \partial_\theta \mathcal{P} \rangle$ and an analogous term for the Hessian. This leads us to kernelized CRFs whose minimizer is given by Eq. (16).

Moreover, due to Corollary 8 we can decompose the solution into a linear combination of vectors $(0, \ldots, 0, \Phi_C((x,y)_C), 0, \ldots, 0)$. With some abuse of notation we identify the latter with $\Phi_C(x, y)$ directly. Clearly $\langle \Phi_C(\cdot), \Phi_{C'}(\cdot) \rangle = 0$ if $C \neq C'$. This yields

$$\langle \Phi_C(x,y), \partial_\theta \mathcal{P} \rangle = \sum_{i=1}^m -k_C((x,y)_C, (x^i, y^i)_C) + \quad (23)$$
$$\sum_{i=1}^m \mathbf{E}_{\bar{y}}\left[k_C((x,y)_C, (x^i, \bar{y})_C)|\theta, x^i\right] +$$
$$\sigma^{-2} \sum_j \alpha_C^j k_C((x,y)_C, (\tilde{x}_C^j, \tilde{y}_C^j))$$

where $\tilde{y}_C^j$ represent all possible instantiations of the $C$ clique of $\tilde{y}^j$ and $\alpha_C^j$ are the expansion coefficients for $\theta$ pertaining to the subspace / clique $C$, for the vector $\Phi_C(\tilde{x}_C^j, \tilde{y}_C^j)$. We should note that, one can also pick a subset of labels on the clique $C$, when using sparse greedy methods, as we describe below.

The projections of the Hessian are given by

$$\Phi_C(x,y)^\top \partial_\theta^2 \mathcal{P} \Phi_{C'}(x',y') \qquad (24)$$
$$= \frac{\delta_{C,C'}}{\sigma^2} k_C((x,y)_C, (x',y')_C) +$$
$$\sum_{i=1}^m \mathrm{Cov}_{\tilde{y}}\left[k_C((x,y),(x^i,\bar{y})), k_{C'}((x',y'),(x^i,\bar{y}))|x^i\right]$$

Since the negative log-posterior is a convex function, (24) will lead to a positive semidefinite matrix, when evaluating the Hessian for different vectors $\Phi_C(x,y)$. Unfortunately the latter of the two terms, namely the sum of covariances, may be expensive to compute, as it involves correlations between labels in different cliques.

Two methods exist to alleviate this problem: firstly we can simply optimize over one subspace at a time. This corresponds to a conditional maximum-a-posteriori estimate per clique. In terms of optimization this is commonly known as subspace descent.

A second method is to approximate the Hessian by a block-diagonal matrix which has entries only for matching cliques, i.e. $C = C'$. In this case, Newton's method turns into a block-preconditioned gradient descent, also known as a block-Jacobi method. Effectively we perform subspace descent on all subspaces simultaneously and re-compute the principal blocks of the Hessian each time.

### 5.2 Subspace Optimization

We will now make specific assumptions about the parameterization of the CRFs under consideration. For instance in the case of Markovian chain structure, e.g. for sequence annotation, we can exploit stationarity to reduce the dimensionality of the parameter space by assuming that all cliques share the same potential function. This leads to a coupling of terms between the individual cliques. In short all $\theta_c$ on corresponding cliques match: in the problem below the matching cliques would be both $(y_t, y_{t+1})$ and $(x_t, y_t)$ for all $t$.

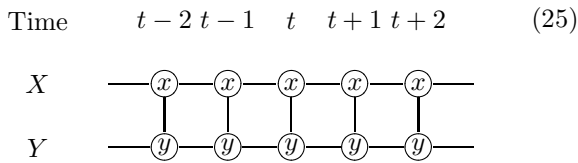

$$\text{Time} \quad t-2 \; t-1 \quad t \quad t+1 \; t+2 \qquad (25)$$

In the above case this leads to

$$\theta = \begin{bmatrix} \mathbf{1} & 0 & \mathbf{1} & 0 & \ldots \\ 0 & \mathbf{1} & 0 & \mathbf{1} & \ldots \end{bmatrix}^\top \begin{bmatrix} \theta_{xy} \\ \theta_y \end{bmatrix}. \qquad (26)$$

Combination of stationarity with the factorization leads to the following expansion for $\theta$:

$$\theta = \sum_l \alpha_l \tilde{\Phi}_{\mathcal{C}_l}(x_l, y_l) \text{ where} \qquad (27)$$
$$\tilde{\Phi}_{\mathcal{C}_l}(x_l, y_l) = (\Phi_{C_l}(x_l, y_l), 0, \Phi_{C_l}(x_l, y_l), 0, \ldots)$$

That is, $\tilde{\Phi}_{\mathcal{C}_l}$ arises from replicating the sufficient statistics of the clique for every position for which the natural parameters are tied. Here $\mathcal{C}_l$ denotes a set of cliques with identical natural parameters (e.g. all $(y_t, y_{t+1})$ pairs). We define formally $\tilde{\Phi}_{\mathcal{C}_l}(x,y)$ to be the vector formed by replicating the same $\Phi_C(x,y)$ for all $C \in \mathcal{C}_l$ and 0 otherwise.[1] We can now put this to practical use and compute the projected gradients and the Hessian, as they arise from (23) and (24).

$$\left\langle \tilde{\Phi}_{\mathcal{C}_j}(x_j, y_j), \partial_\theta \mathcal{P} \right\rangle \qquad (28)$$
$$= \sum_{i=1}^m \sum_{C \in \mathcal{C}_j} -k_C((x_j, y_j), (x^i, y^i)_C) + \qquad (29)$$
$$\sum_{i=1}^m \sum_{C \in \mathcal{C}_j} \mathbf{E}_{\bar{y}}\left[k_C((x_j, y_j),(x^i, \bar{y})_C)|\theta, x^i\right]$$
$$\sigma^{-2} \sum_{l, C \in \mathcal{C}_l} \alpha_l k_C((x_j, y_j),(x_l, y_l))$$

where we assumed that $\theta = \sum_l \alpha_l \tilde{\Phi}_{\mathcal{C}_l}(x_l, y_l)$ for suitably chosen $(x_l, y_l)$ pairs. The Hessian could be computed in the same fashion. However, as this involves computing long-range correlations between various cliques, we use the block-Jacobi approximation of (24). Hence for clique sets $\mathcal{C}_i = \mathcal{C}_j$ we have

$$\tilde{\Phi}_{\mathcal{C}_l}^\top(x_l, y_l) \left[\partial_\theta^2 \mathcal{P}\right] \tilde{\Phi}_{\mathcal{C}_j}^\top(x_j, y_j) \approx \qquad (30)$$
$$\sigma^{-2} k_C((x_l, y_l),(x_j, y_j)) +$$
$$\sum_{i=1, C \in \mathcal{C}_l}^m \mathrm{Cov}_y\left[k_C((x_l,y_l),(x^i,y)_C), k_C((x_j,y_j),(x^i,y)_C)\right]$$

In other words, we are ignoring correlations which go beyond the clique boundaries. The price we pay for this is slower convergence. Worst case, the convergence will slow down from quadratic to linear, as the optimization will behave like a subspace Gauss-Southwell method.

### 5.3 Sparse Greedy Approximation

The number of parameters $\alpha_i$ is still enormous: for an optimal solution we need as many coefficients as there are $(x^i, y)_C$ restrictions (with $y \in \mathcal{Y}$) on the cliques. For instance in the sequence annotation model, the number of coefficients required would still scale with the length of the total strings $x$. As a further reduction in dimensionality we use sparse greedy approximations along the lines of [Smola and Schölkopf, 2000, Fine and Scheinberg, 2001].

---

[1] Clearly this implies that all cliques in $\mathcal{C}_l$ have the same functional form for the sufficient statistics. Note that there is no need to assume that *all* cliques with compatible sufficient statistics have the same potential function.

The main (and only) difference is that now we perform the decompositions not on $k((x, y), (x', y'))$ directly. Instead we use low-rank approximations for each of the subspaces spanned by $\Phi_C(x, y)$ directly. The reason is that by doing so we can capture a much more representative subspace, as pointed out in Corollary 8.

An adaptation of the selection strategy of Fine and Scheinberg [2001] for a good set of $\Phi_C(x, y)$ in the case of stationarity is equivalent to performing an incomplete Cholesky factorization on the matrix of scalar products $k_C((x, y), (x', y'))$ between all matching cliques $C \in \mathcal{C}_l$. The pivots are then used as basis vectors. The pivots are selected to be the vectors with the currently largest residuals. This method has $O(mn)$ time-complexity per factor, where $m$ is the number of candidate functions and $n$ is the number of dimensions chosen so far. See [Schölkopf and Smola, 2002] for further implementation details.

## 6 Experiments

We now proceed to experiments on a specific task, namely pitch accent prediction. Pitch accent prediction, a sub-task of speech recognition, is detecting the words that are more prominent than others in an utterance. We model this problem as a sequence annotation problem as in (25), where $\mathcal{Y} = \{\pm 1\}^T$, that is, we have a sequence of binary labels of length $T$.

Note that the clique-structure of (25) comprises of sets $(y_t, y_{t+1})$, that is, adjacent labels, and sets $(x_t, y_t)$, that is, labels plus a local neighborhood of data. Regarding the cliques $(x_t, y_t)$ the following lemma may be used:

**Lemma 13 (Centering of Labels)** *The transformation $\Phi(x, y) \leftarrow \Phi(x, y) - \mu$ with $\mu = |\mathcal{Y}|^{-1} \sum_{y \in \mathcal{Y}} \Phi(x, y)$ leaves the conditional probabilities $p(y|x; \theta)$ unchanged.*

**Proof** Direct calculation: all this transformation does is to change all exponential terms by a constant, which is absorbed in the log-partition function. ∎

A consequence is that $\Phi_C(x_t, y_t) = y_t \Phi_C(x_t)$, which means that the kernel is given by $y_t y'_t k(x_t, x'_t)$.

Given Eq. (28) and Eq. (30), the remaining challenge is to come up with an efficient way of comput-ing the expectations and covariances. Due to Lemma 13, these computations reduces to the computation of $\mathbf{E}_y[y_C|\theta, x^i]$, which are calculated for every clique of every training sequence by the standard forward-backward algorithm using transition probability matrix and the observation probability matrix defined with respect to $\theta$.

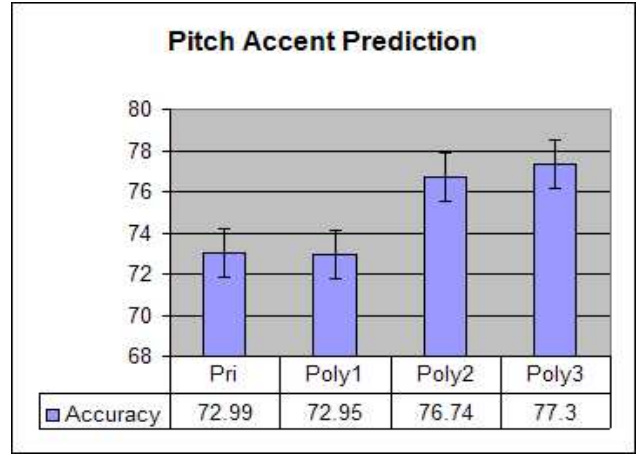

Figure 1: Test accuracy of pitch accent prediction task over a window of size 5 using 5-fold cross validation.

We used Switchboard Corpus [Godfrey et al., 1992] to experimentally evaluate the described method. The data consists of adult telephone conversations and was phonetically transcribed and annotated. We extracted 500 sentences from this corpus and ran experiments using 5-fold cross validation. Features consist of simple binary features as well as real valued features extracted over a window of size 5 centered at the current position in the sequence. Details of the feature representation can be found in Gregory and Altun [2004]. Since most of the features are binary, we choose polynomial kernel of different degrees for kernelization of the CRFs.

We ran experiments to optimize polynomial degree 1, 2 and 3 kernels using the block-Jacobi method. We also ran experiments where the polynomial degree 1 CRF was optimized using the more standard primal optimization as described in Sha and Pereira [2003], Gregory and Altun [2004] (labeled *Pri* in the figure). The results are reported in Figure 1. Not surprisingly, the primal and dual optimizations of polynomial degree 1 are (almost) the same. Using second and third degree features increase the performance substantially.

## 7 Discussion

The application of decomposition results in Section 3 to CRFs only scratch the surface of what is possible with nonparametric graphical models. Clearly, we can use the same set of ideas for conventional undirected graphs, which opens a large toolbox to kernel methods.

Some of the obvious applications are estimation in the presence of missing values and clustering. One of the main technical difficulties to be overcome in this context are the common problem of being unable to compute log-partition functions exactly. Here approximate results, such as tree-decompositions are needed. Other means to alleviate this problem are MCMC and exact sampling approaches.

Even if the above problems are overcome, the resulting nonconvex optimization problems still present a formidable challenge. Here semidefinite relaxations of the original problem may be useful. This is the subject of ongoing research.

**Acknowledgments** National ICT Australia is funded through the Australian Government's *Backing Australia's Ability* initiative, in part through the Australian Research Council. This work was supported by grants of the ARC and NSF-ITR grants IIS-0312401 and IIS-0085940. Thanks to Michelle Gregory for providing us the pitch accent data and the valuable features.